\pdfoutput=1

\documentclass[11pt]{article}

\usepackage[]{acl}

\usepackage{times}
\usepackage{latexsym}
\usepackage{amsmath}
\usepackage{amssymb}
\usepackage{booktabs}
\usepackage{array, makecell} 
\usepackage{epsfig}
\usepackage{graphicx}
\usepackage{subcaption}
\usepackage{caption}
\usepackage{tikz}
\usepackage{pgfplots}
\usepackage{enumitem} 
\usetikzlibrary{fit}
\pgfplotsset{compat=newest}%

\usepackage[T1]{fontenc}

\usepackage[utf8]{inputenc}

\usepackage{microtype}

\usepackage{inconsolata}

\usepackage{graphicx}
\usepackage{subcaption}

\usepackage{algorithm}
\usepackage{algpseudocode}

\usepackage{overpic}   
\usepackage{fancyvrb}

\title{LLMVoX: Autoregressive Streaming Text-to-Speech Model for Any LLM}

\author{
Sambal Shikhar$^{1}$ \quad Mohammed Irfan Kurpath$^{1}$ \quad Sahal Shaji Mullappilly$^{1}$ \quad Jean Lahoud$^{1}$ \\
\textbf{Fahad Khan}$^{1,2}$ \quad \textbf{Rao Muhammad Anwer}$^{1}$ \quad \textbf{Salman Khan}$^{1}$ \quad \textbf{Hisham Cholakkal}$^{1}$ \\
$^1$Mohamed Bin Zayed University of Artificial Intelligence (MBZUAI), UAE\\
$^2$Linköping University, Sweden\\
\texttt{\{sambal.shikar, hisham.cholakkal\}@mbzuai.ac.ae}
}

\begin{document}
\maketitle
\begin{abstract}
Recent advancements in speech-to-speech dialogue systems leverage LLMs for multimodal interactions, yet they remain hindered by fine-tuning requirements, high computational overhead, and text-speech misalignment. Existing \textit{speech-enabled LLMs} often degrade conversational quality by modifying the LLM, thereby compromising its linguistic capabilities.
In contrast, we propose \textbf{LLMVoX}, a \textbf{\textit{lightweight 30M-parameter}}, \textit{LLM-agnostic, autoregressive streaming TTS} system that generates high-quality speech with low latency, while fully preserving the capabilities of the base LLM. Our approach achieves a significantly lower Word Error Rate compared to speech-enabled LLMs, while operating at comparable latency and UTMOS score.
By decoupling speech synthesis from LLM processing via a multi-queue token streaming system, LLMVoX supports  seamless, infinite-length dialogues.  Its \textit{plug-and-play} design also facilitates extension to various tasks with different backbones. Furthermore, LLMVoX generalizes to new languages with only dataset adaptation, attaining a low Character Error Rate on an Arabic speech task. 
 Additionally, we have integrated LLMVoX with a Vision-Language Model to create an omni-model with speech, text, and vision capabilities,  without requiring additional multimodal training. Our code base and project page is  available at \href{https://mbzuai-oryx.github.io/LLMVoX/}{mbzuai-oryx.github.io/LLMVoX}
\end{abstract}

\section{Introduction}
Large Language Models (LLMs) have excelled in the new era of conversational AI, transforming how machines understand, generate, and interact with humans. While most LLMs were initially designed for text-based interactions, there are some recent efforts toward more natural and intuitive \emph{speech-to-speech} dialogue systems, allowing users to engage with AI models through spoken language. 

\begin{figure}[t]
    \centering
    \begin{overpic}[width=0.49\textwidth,trim={0 2mm 0 0},clip]{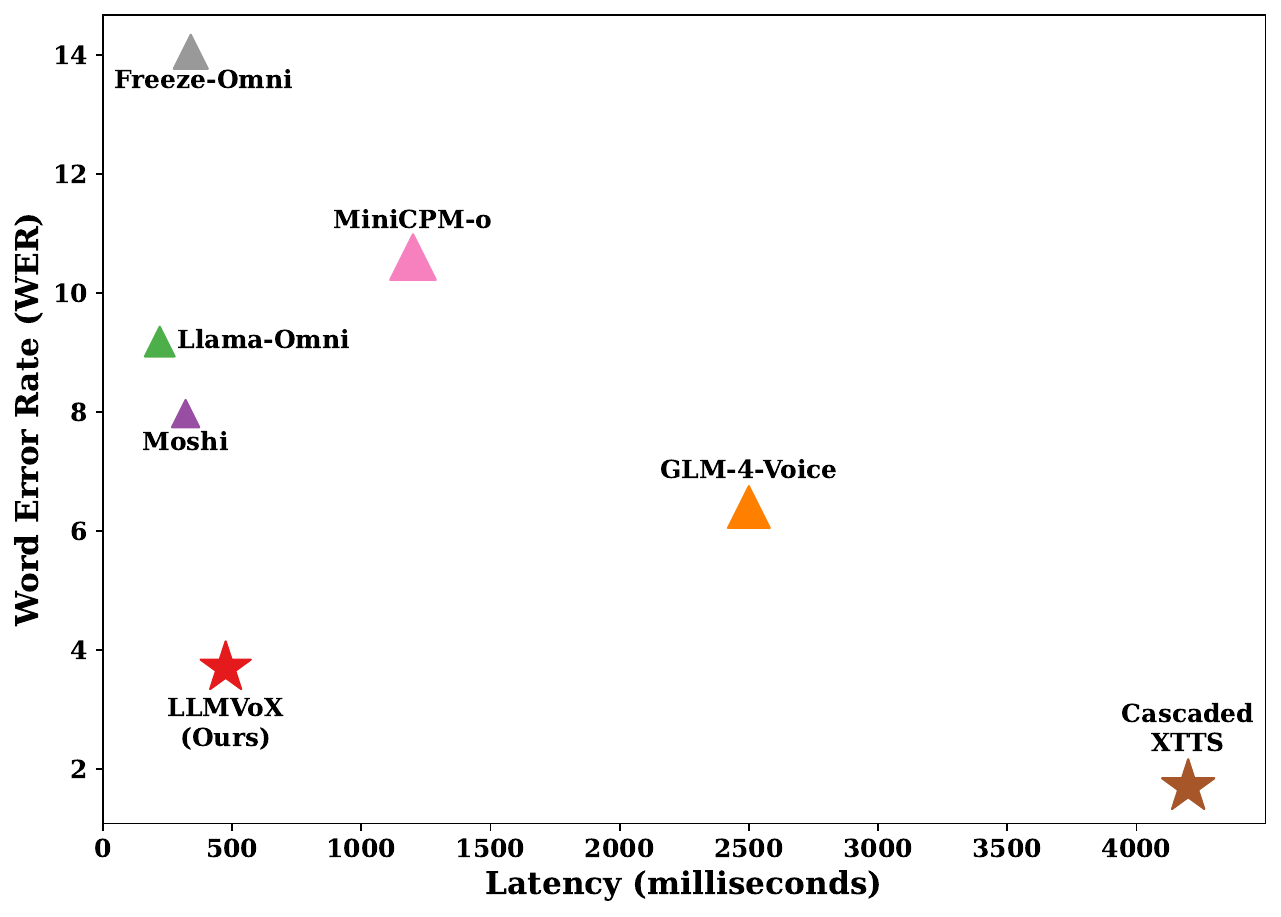}
        \put(45, 53){ 
        \setlength{\tabcolsep}{3pt}
        \scalebox{0.55}{
            \begin{tabular}{l|c|c}
Model & GPT Score ($\uparrow$) & WER ($\downarrow$) \\
\midrule
Cascaded XTTS  & 7.20  & 1.70 \\
Llama-Omni  & 3.64  & 9.18  \\
Moshi  & 3.31  & 7.97  \\
GLM-4-Voice  & 5.30  & 6.40  \\
Freeze-Omni  & 4.23  & 14.05  \\
MiniCPM-o 2.6  & 5.84  & 10.60  \\
\midrule
\textbf{LLMVoX (Ours)}  & \textbf{6.88}  & \textbf{3.70}  \\
\bottomrule
\end{tabular}
        }
        }
    \end{overpic}
    \caption{
    Speech quality (WER) vs latency (milliseconds) comparison of recent speech-enabled LLMs.  
   Our LLMVoX is  LLM-agnostic streaming TTS  that generates high-quality speech (lower WER) comparable to XTTS \cite{casanova2024xtts} while operating  $~10$× faster.  In the plot, $\triangle$ represents LLM-dependent methods, and $\bigstar$ denotes LLM-agnostic methods. The size of each symbol is proportional to the GPT score, indicating overall response quality. All methods are evaluated under similar settings and use similarly sized base LLMs.}
    \label{fig:teaser}
\end{figure}

Existing speech-enabled LLMs typically aims to \textit{unify text and speech processing} within a single, fine-tuned LLM. Recent models such as Kyōtai Moshi \cite{defossez2024moshi}, Mini-Omni~\cite{xie2024mini}, LLaMA-Omni~\cite{fang2024llama}, and Freeze-Omni~\cite{wang2024freeze} extend or modify pretrained text-based LLMs, enabling them to directly handle spoken inputs and outputs. Although these end-to-end systems can offer faster and streamlined speech generation, they require \textit{large-scale fine-tuning} of LLM on multimodal data. This fine-tuning with speech data often compromises the original reasoning and expressive capabilities of the base LLM \cite{chen2024voicebench,defossez2024moshi,kalajdzievski2024scaling,zhai2023investigating}, while also imposing substantial computational and data requirements for speech adaptation.  Moreover, these architectures often condition speech adaptation on LLM hidden states, making them inherently \textit{LLM-dependent}, thereby requiring re-adaptation for each base LLM. 

Alternatively, an \textit{LLM-agnostic} approach is to leverage a \textit{\textbf{cascaded pipeline}},
where speech is converted to text via automatic speech recognition (ASR), processed by an LLM to generate a textual response, and finally passed through a text-to-speech (TTS) module for speech output. This cascaded approach offers several advantages, including the availability of diverse off-the-shelf ASR \cite{radford2023robust}, LLM \cite{fang2024llama}, and TTS \cite{casanova2024xtts} models, the preservation of base LLM capabilities, improved speech quality, and an \textit{LLM-agnostic} design that allows seamless adaptation to any base LLM in a plug-and-play manner, without the need for computationally expensive model retraining.  However, such cascaded approaches often introduce high latency (see Cascaded-XTTS in Figure~\ref{fig:teaser}), making real-time interactions challenging. The primary reason for this high latency is the incompatibility between the autoregressive nature of LLM-based text generation and conventional TTS models, which typically process text inputs collectively, despite the text being available incrementally from LLM. This prevents speech generation from starting until the entire text response, or a large chunk of it, has been generated by the LLM. Furthermore, many existing TTS models rely on non-streaming speech decoders, leading to a larger delay between text and speech \mbox{generation.}  

To address the aforementioned limitations of existing speech-enabled LLMs, we propose \textbf{LLMVoX}, an  \emph{autoregressive}, \emph{LLM-agnostic} streaming framework. It aims to \emph{preserve the underlying LLM’s capabilities} by completely decoupling speech synthesis from the LLM, while enabling high-quality, low-latency speech generation (Figure \ref{fig:teaser}) in an autoregressive setting, \textit{running in parallel with the LLM’s text generation}.

\subsection{Contributions}
Our LLMVoX leverages a lightweight transformer \cite{waswani2017attention} to generate \emph{discretized speech tokens} in an autoregressive manner from streaming LLM text, making it straightforward to “plug” into any existing LLM pipeline without model retraining or fine-tuning. 

LLMVoX adopts a multi-queue streaming approach to enable continuous and potentially \textit{infinite-length} speech generation. By maintaining acoustic continuity and avoiding awkward pauses during extended dialogues, this design helps sustain a fluid user experience with minimal latency of 475 milliseconds for the entire cascaded pipeline including ASR  \cite{radford2023robust}, LLaMA-3.1-8B \cite{fang2024llama}, and LLMVoX (Figure \ref{fig:teaser}). 

Furthermore, we demonstrate the generalization ability of the LLMVoX architecture to languages other than English by \textit{\textbf{adapting it to Arabic}} for seamless plugging with Arabic LLM like Jais \cite{sengupta2023jais}.
This adaptation requires only a simple change in the LLMVoX training data to Arabic, without any architectural modifications, such as explicit Grapheme-to-Phoneme (G2P) conversion \cite{nguyen2023xphonebert,cherifi2021arabic,jung2006grapheme}, and can be similarly applied to any new language. Moreover, we integrated LLMVoX with a Vision Language Model (VLM) \textbf{\textit{to obtain an omni-model}} with speech, text, and vision capabilities without explicit multimodal training. 

\textit{\textbf{The key  contributions of our method \mbox{are summarized below}}:}\\
\noindent\textbf{(i)} We introduce LLMVoX, \textbf{\textit{a lightweight 30M-parameter, LLM-agnostic,  autoregressive streaming TTS}} framework that offers a plug-and-play solution for seamless integration with any off-the-shelf LLM or VLM—without fine-tuning or architectural modifications.

\noindent\textbf{(ii)} We use a \textit{\textbf{multi-queue streaming mechanism} }that enables continuous, low-latency speech generation and \emph{infinite-length speech}, effectively adapting to LLMs with different context lengths.\\
\noindent\textbf{(iii)} Our comprehensive experiments demonstrate that \textbf{\textit{LLMVoX performs favorably compared to state-of-the-art speech-enabled LLMs}} in speech quality and latency while preserving the underlying LLM capabilities. Our cascaded system with LLMVoX achieves a WER of 3.70, maintains high speech quality with a UTMOS of 4.05, and delivers an end-to-end latency of 475ms (see Figure~\ref{fig:teaser}).\\ 
\noindent\textbf{(iv)} We demonstrate LLMVoX's \textbf{\textit{ability to generalize to other languages, such as Arabic}}, by simply modifying the training data-without any architectural changes. To this end, \textbf{\textit{we generated 1,500 hours (450k pairs) of a synthetic, single-speaker Arabic text-speech dataset }}.  

\noindent \textbf{(v)} Adapting LLMVoX to Arabic results in \textbf{\textit{the first streaming, autoregressive Arabic speech generator that can be seamlessly integrated with any Arabic LLM}}, such as Jais \cite{sengupta2023jais}, to create Arabic speech-enabled LLMs. LLMVoX achieves a \textbf{CER} of $\sim8\%$ comparable to even non-streaming Arabic TTS methods, while operating at lower latency—demonstrating the scalability and adaptability of our approach.\\
\noindent\textbf{(vi)} We further \textbf{\textit{integrate LLMVoX with QWen 2.5-VL-7B VLM ~\cite{qwen2.5-VL} to obtain an omni-model with speech, text, and vision capabilities}} that do not require explicit multimodal training. This model \textbf{\textit{performs favorably when compared to the state-of-the-art omni-model}}, MiniCPM-o 2.6 \cite{minicpm}, in visual \textit{speech} question answering on  LLaVA-Bench (in the wild) \cite{liu2024mmbench}, while achieving 30\% lower latency.

\section{Related Work}
\label{sec:related_work}

Here, we review recent speech-enabled LLMs, followed by various speech tokenization methods employed in TTS models and speech-enabled LLMs. \vspace{0.05cm}\\

\noindent \textbf{Speech-enabled LLMs:} Models such as Qwen-2 Audio~\cite{chu2024qwen2}, VITA~\cite{fu2024vita}, Ichigo~\cite{dao2024ichigo}, and Baichuan-Omni~\cite{li2024baichuan} append speech adapters to LLMs for speech-to-text tasks, yet still rely on separate TTS modules, inheriting latency issues from cascaded pipelines. SpeechGPT~\cite{zhang2023speechgpt}, AudioPaLM~\cite{rubenstein2023audiopalm}, EMOVA~\cite{chen2024emova}, and AnyGPT~\cite{zhan2024anygpt} integrate speech tokens directly into LLM vocabularies for end-to-end multimodal inference; however, as chain-of-modality methods, they incur latency by waiting for the complete text response before speech generation. Recent speech-enabled LLMs targeting low-latency interactions include Kyōtai Moshi~\cite{defossez2024moshi}, which employs a dual-channel architecture with Mimi Neural Audio Codec for real-time dialogue; Mini-Omni~\cite{xie2024mini}, which combines text and speech modeling with batch-parallel inference to reduce delays; and LLaMA-Omni~\cite{fang2024llama}, which uses a CTC-based mechanism (latency $\sim$236ms). GLM-4-Voice~\cite{zeng2024glm} trains on a trillion bilingual tokens with a low-bitrate (175bps) tokenizer for high-fidelity synthesis at higher compute cost; MiniCPM-o 2.6~\cite{minicpm,yao2024minicpm} adopts an omni-modal LLM with a streaming speech decoder for real-time synthesis. Closer to our approach, Freeze-Omni~\cite{wang2024freeze} mitigates catastrophic forgetting by freezing the base LLM and integrating speech-specific modules. They employ a 3 stage training where LLM parameters are kept frozen throughout but in the final stage of training, Freeze-Omni conditions its speech decoder on LLM hidden states, necessitating retraining the speech components for any new base LLM, thereby limiting its plug-and-play capability.
\noindent \textbf{Speech Tokenization:} Mapping waveforms to discrete tokens compatible with transformers has advanced speech-to-speech modeling. Neural acoustic codecs (e.g., EnCodec~\cite{defossez2022high}, LauraGPT~\cite{du2023lauragpt}) employ residual vector quantization (RVQ) for high-fidelity synthesis; hybrid approaches (e.g., SpeechTokenizer~\cite{zhang2023speechtokenizer}) use hierarchical RVQ layers to enhance phonetic representation; and supervised tokenizers (e.g., CosyVoice~\cite{du2024cosyvoice}) integrate vector quantization into ASR for improved text-speech alignment. Mimi~\cite{defossez2024moshi} employs split-RVQ for balanced phonetic \mbox{discrimination and quality.}
\section{Methodology}
\label{sec:methodology}

\begin{figure*}
\centering
\includegraphics[width=0.9\textwidth]{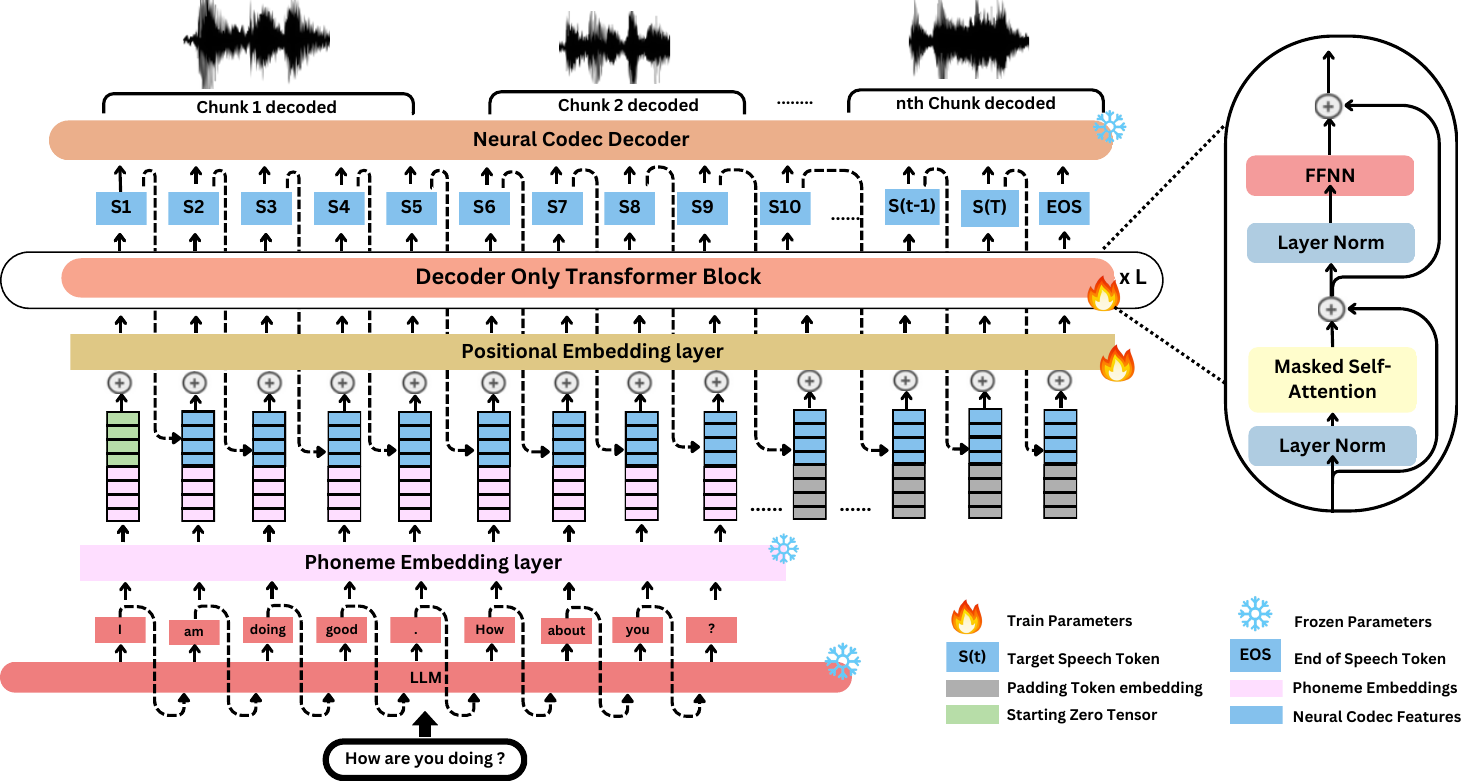}
\caption{Overview of the proposed architecture. Text from the LLM is tokenized via a ByT5-based Grapheme-to-Phoneme(G2P) model, producing byte-level phoneme embeddings (teal). These are concatenated with the previous speech token’s feature vector (blue), L2-normalized, and fed into a decoder-only Transformer to generate the next token. A neural codec (WavTokenizer) decoder (orange) reconstructs speech every n speech tokens predicted.}
\label{fig:architecture}
\end{figure*}
Our proposed LLMVoX system in Figure~\ref{fig:architecture} is a fully autoregressive Text-to-Speech (TTS) framework designed to convert text outputs from an upstream Large Language Model (LLM) into high-fidelity streaming speech. The central motivation behind our design is to decouple the speech synthesis component from the text-generation process so that the inherent reasoning and expressive capabilities of the LLM remain unaltered while not compromising latency. By recasting TTS as a token prediction task over discrete acoustic units, we leverage Transformers architecture \cite{waswani2017attention} and neural audio representations to achieve natural, low-latency speech generation.

In our approach, the speech signal is represented as a sequence of discrete tokens drawn from a fixed vocabulary of 4096 entries. These tokens are generated by a neural audio codec, and the speech token is predicted token-by-token in an autoregressive manner. Figure~\ref{fig:architecture} provides an overview of the overall architecture, where phoneme-aware embeddings derived from Grapheme-to-Phoneme (G2P)~\cite{zhang2022byt5} model are combined with previous acoustic context and processed by a decoder-only Transformer to predict the next speech token.

\subsection{Neural Audio Tokenization}
\label{sec:codec}
To model speech generation as an autoregressive task using Transformers \cite{wang2023neural}, we use a neural audio codec that discretizes the continuous audio waveform using a single-layer residual vector quantization (RVQ) such as \textbf{WavTokenizer}~\cite{ji2024WavTokenizer}. WavTokenizer yields a compact representation that supports high-quality speech reconstruction while keeping sequence lengths manageable. Given a 24\,kHz waveform \(\mathbf{x}\), the encoder \(\mathrm{Enc}(\cdot)\) extracts latent feature vectors \(\{\mathbf{f}_1, \mathbf{f}_2, \dots, \mathbf{f}_T\}\), where \(T\) is the number of tokens. Each feature \(\mathbf{f}_t\) is quantized via \(S_t = \mathrm{VQ}(\mathbf{f}_t)\) with \(S_t \in \{1, \dots, 4096\}\). Typically, 40--75 tokens represent one second of speech. The decoder \(\mathrm{Dec}(\cdot)\) then reconstructs the audio waveform from these discrete token indices.

\subsection{Byte-Level Grapheme-to-Phoneme Embedding}
\label{sec:g2p}

To infuse phonetic information into the synthesis process without incurring the overhead of explicit phoneme prediction, we employ the embedding layer of a ByT5-based Grapheme-to-Phoneme (G2P) model~\cite{zhang2022byt5}. This decision is driven by two main considerations:
(1) \textit{Phonetic Richness}: This ByT5 based G2P model is fine-tuned on over 100 languages, so its embeddings capture subtle phonetic similarities and distinctions, ensuring accurate pronunciation, and (2) \textit{Computational Efficiency:} By directly reusing the learned embeddings as a “table lookup”, we avoid extra computation needed for explicit phoneme conversion, thus reducing latency.

\paragraph{Embedding Extraction and Padding Alignment.}  
Let \(\tilde{t}_1, \tilde{t}_2, \dots, \tilde{t}_N\) denote the sequence of words produced by the LLM. Each word \(\tilde{t}_i\) is decomposed into byte-level sub-tokens using the ByT5 tokenizer, i.e., \(\tilde{t}_i \rightarrow [\beta^i_1, \beta^i_2, \dots, \beta^i_{n_i}]\), where \(n_i\) is the number of sub-tokens for token \(\tilde{t}_i\). Let \(M\) be the total number of sub-tokens from all text tokens. Each sub-token \(\beta^i_j\) is then mapped to an embedding vector as \(\mathbf{b}^i_j = \mathrm{Embed}_{\mathrm{ByT5}}(\beta^i_j)\), \mbox{where \(\mathbf{b}^i_j \in \mathbb{R}^{256}\).}

The ground-truth speech is tokenized into a sequence of \(T\) discrete speech tokens using WavTokenizer\cite{ji2024WavTokenizer}, where typically \(T > M\). To align the length mismatch we pad the sub-token sequence to length \(T\). Formally, the padded text embedding sequence \(\{\mathbf{b}_1, \mathbf{b}_2, \dots, \mathbf{b}_T\}\) \mbox{is defined as:}
\[
\mathbf{b}_t = 
\begin{cases}
\mathrm{Embed}_{\mathrm{ByT5}}(\beta_t), & \text{if } 1 \leq t \leq M, \\
\mathbf{b}_{\mathrm{PAD}}, & \text{if } M < t \leq T,
\end{cases}
\]
where \(\beta_t\) is the \(t\)-th sub-token and \(\mathbf{b}_{\mathrm{PAD}} \in \mathbb{R}^{256}\) is the embedding for the \texttt{<PAD>} token (obtained from the ByT5 embedding layer)\cite{xue2022byt5}. Although \(\mathbf{b}_{\mathrm{PAD}}\) does not encode phonetic information, the Transformer's self-attention mechanism will use context from the previous inputs \mbox{to refine its representation.}

\subsection{Input Representation} \label{sec:input_representation}  
At each time step \(t\) (\(t=1,\dots,T\)), the input vector is constructed by concatenating the phoneme embedding \(\mathbf{b}_t \in \mathbb{R}^{256}\) with the latent acoustic feature vector \(\mathbf{f}_{t-1} \in \mathbb{R}^{512}\) from the previous speech token \(S_{t-1}\), forming \(\mathbf{x}_t = [\mathbf{b}_t;\mathbf{f}_{t-1}] \in \mathbb{R}^{768}\). This vector is L2-normalized, and a learnable positional embedding \(\mathbf{r}_t \in \mathbb{R}^{768}\) is added, yielding \(\mathbf{z}_t = \mathbf{x}_t + \mathbf{r}_t\). The sequence \(\{\mathbf{z}_1,\mathbf{z}_2,\dots,\mathbf{z}_T\}\) is then fed into the decoder-only Transformer as shown in Figure~\ref{fig:architecture}.

\subsection{Decoder-Only Transformer for Speech Token Generation} \label{sec:autoregressive}  
The core of our synthesis model is a lightweight decoder-only Transformer (4 layers) that autoregressively predicts the sequence of speech tokens \(S_1, S_2, \dots, S_T\). Our objective is to model the conditional probability \( p\bigl(S_t \mid S_1, S_2, \dots, S_{t-1}, \{\mathbf{z}_1, \mathbf{z}_2, \dots, \mathbf{z}_T\}, \theta\bigr) \) for each \(t = 1, \dots, T\), where \(\theta\) denotes the transformer's. Moreover,
At \(t=1\), no previous speech token is available. We thus initialize the acoustic context with a zero tensor ensuring that the model receives a consistent starting signal.
\subsection{Training Objective and Procedure}
\label{sec:training}

Training LLMVoX involves minimizing the cross entropy loss over the ground-truth speech token sequence \(\{S_1, \dots, S_T\}\):
\begingroup
\setlength{\abovedisplayskip}{3pt}
\setlength{\belowdisplayskip}{3pt}
\[
\mathcal{L} = -\sum_{t=1}^{T} \log p\bigl(S_t \mid S_{<t}, \mathbf{z}, \theta\bigr).
\]
\endgroup
A causal mask is applied within the Transformer to enforce the autoregressive property.

\begin{algorithm}[t]
\small
\caption{Streaming Inference with Adaptive Chunk Size (Parallel Text Generation)}
\label{alg:streaming_inference_parallel}
\begin{algorithmic}[1]
\Require Speech query \(\mathbf{x}_{\text{user}}\)
\Ensure Real-time speech \(\hat{\mathbf{x}}\)
\State \(ASR\text{-}Text \gets \mathrm{ASR}(\mathbf{x}_{\text{user}})\)
\State \(\mathrm{LLM\text{-}Text} \gets \mathrm{LLM}(ASR\text{-}Text)\) \quad // Generate text tokens in parallel
\State Enqueue generated text tokens into FIFO queue \(\mathcal{Q}_0\)
\State Split \(\mathcal{Q}_0\) into FIFO queues \(\mathcal{Q}_1\) and \(\mathcal{Q}_2\) (by sentence boundaries)
\ForAll{\(i\in\{1,2\}\) \textbf{in parallel}}
    \State \(\{S_1,\dots,S_M\}\gets\mathrm{LLMVoX}_i(\mathcal{Q}_i)\) \quad // Generate speech tokens
    \State \(chunk\_size \gets n,\; startIdx \gets 1\)
    \While{\(startIdx \le M\) \textbf{and} speech ongoing}
        \State \(endIdx \gets \min(startIdx+chunk\_size-1, M)\)
        \State Decode \(\{S_{startIdx},\dots,S_{endIdx}\}\to\hat{\mathbf{x}}^{(m)}_i\); Enqueue into \(\mathcal{P}_i\)
        \State \(startIdx \gets endIdx+1,\; chunk\_size \gets 2\cdot chunk\_size\)
    \EndWhile
\EndFor
\State \textbf{Stream speech:} Dequeue and stream chunks from \(\mathcal{P}_1\) and \(\mathcal{P}_2\) concurrently until complete.
\end{algorithmic}
\label{algorithm}
\end{algorithm}

\section{Streaming Inference}\label{sec:inference}
\begin{figure}[t]
    \centering
    \includegraphics[width=0.95\columnwidth]{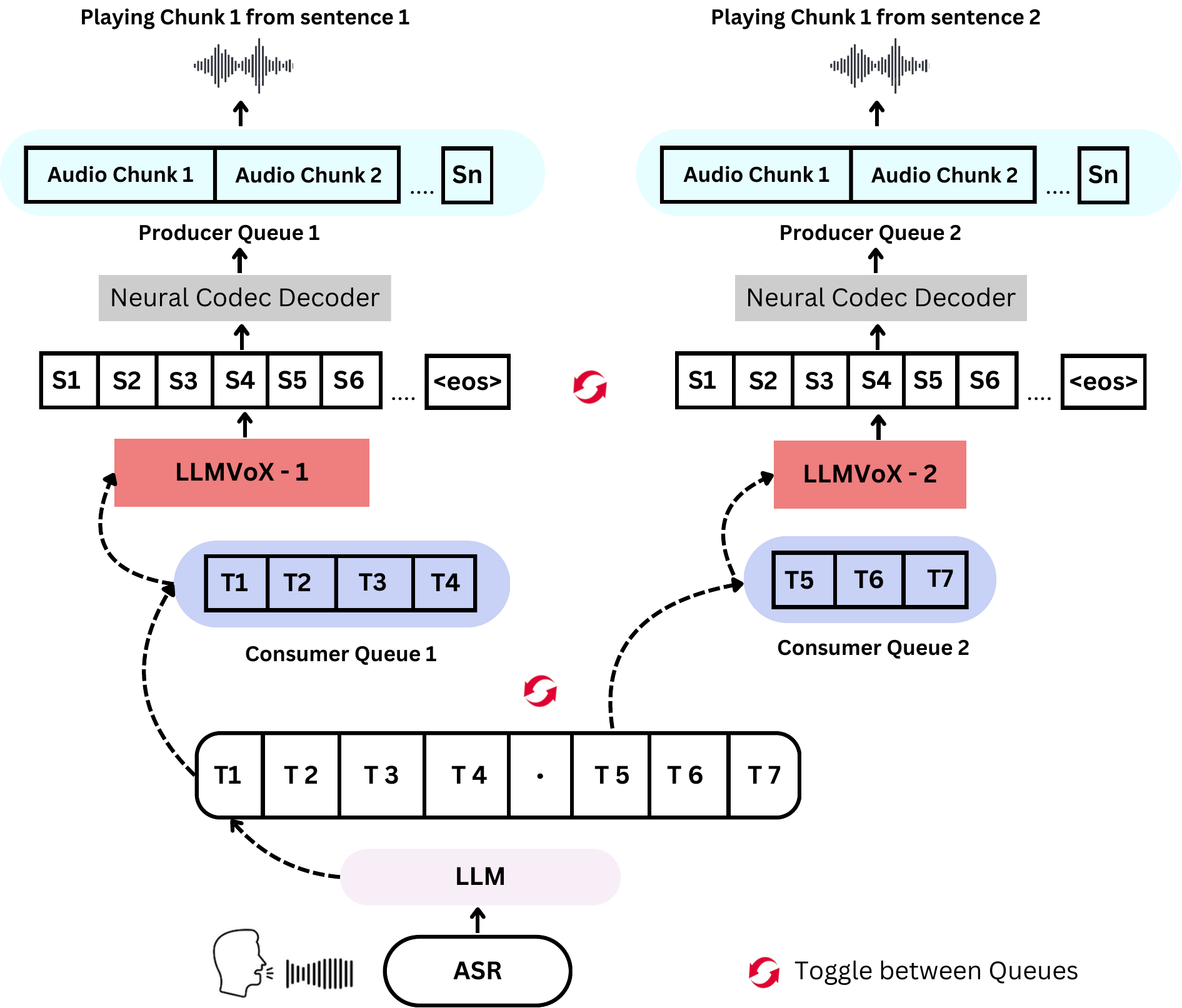}
    \caption{Overview of our streaming inference pipeline. Two replica TTS modules process text in parallel from two queues and place them into two producer queues.}
    \label{fig:streaming_architecture}
\end{figure}

\indent We adopt a low-latency streaming inference pipeline (Figure~\ref{fig:streaming_architecture} and Algorithm \ref{algorithm}) for real-time speech dialogue system. Given the user's speech input \(\mathbf{x}_{\text{user}}\), we first transcribe it using an ASR model (e.g., Whisper) to obtain \(t_{\text{query}} = \mathrm{ASR}(\mathbf{x}_{\text{user}})\). An LLM then generates a stream of words \(\{\tilde{t}_1, \tilde{t}_2, \dots, \tilde{t}_N\} = \mathrm{LLM}(t_{\text{query}})\), which are alternately enqueued into two FIFO queues, \(\mathcal{Q}_1\) and \(\mathcal{Q}_2\), based on sentence boundaries. Two replica TTS modules, \(\mathrm{LLMVoX}_1\) and \(\mathrm{LLMVoX}_2\), concurrently dequeue words from \(\mathcal{Q}_1\) and \(\mathcal{Q}_2\) and predict speech tokens \(\{S_1, S_2, \dots, S_T\} = \mathrm{LLMVoX}_i(\mathcal{Q}_i)\) for \(i \in \{1,2\}\). Every \(n\) speech token generated is then decoded into speech by WavTokenizer decoder and placed in producer queues \(\mathcal{P}_1\) and \(\mathcal{P}_2\) accordingly which is then streamed to the user immediately ensuring uninterrupted playback. The initial chunk size is \(n\) tokens, and after each segment is decoded, the chunk size doubles, leveraging the playback interval of previous speech to allow extra processing time as decoding larger chunks gives better speech output. This toggling mechanism seamlessly handles long or continuous text without requiring models with an extended or \mbox{large \textbf{context window}.}

\section{Experimental Settings}
\label{sec:experimental_settings}
\textbf{Training Dataset}:
We use the \emph{VoiceAssistant-400K} dataset from the Mini-Omni series~\cite{xie2024mini}, which contains over 400K GPT-4o-generated question-answer pairs with corresponding synthesized speech, curated for speech assistant fine-tuning. Our training pipeline uses only the answer text and synthetic speech, resulting in approximately 2,200 hours of single-speaker English speech data. For Arabic, we collected 450K text entries of varying lengths from diverse Hugging Face corpora, cleaned the data, and generated corresponding speech using XTTS~\cite{casanova2024xtts} at a low-temperature setting, yielding about 1,500 hours of single-speaker Arabic speech data.

 \noindent \textbf{Training Configuration}: Our streaming TTS model is a 4-layer, decoder-only Transformer ($n_{\mathrm{embd}}=768$, $n_{\mathrm{head}}=8$) trained with a micro-batch size of 4, \texttt{gradient\_accumulation\_steps} of 8, and a context block size of 8192 tokens. We use AdamW\cite{loshchilov2017fixing} (\texttt{lr}=$3\times10^{-4}$, \texttt{weight\_decay}=0.1) with a 50K-step warmup, then decay the learning rate over 1M steps to $3\times10^{-6}$. Gradients are clipped at a norm of 1.0. The system runs on 4 A100 GPUs for around 3 days, using \texttt{bfloat16} precision. We use \textbf{flash-attention}\cite{dao2022flashattention} for efficient and fast training while also using \textbf{KV-Cache} while inferencing. Under these settings, we separately train English and Arabic models on 2,200 and 1,500 hours of single-speaker speech data, respectively.
\section{Results and Evaluation}
\label{sec:results}
\begin{table*}[ht]
\centering
\scriptsize  
\setlength{\tabcolsep}{3pt} 
\begin{tabular}{lcccccccc}
\toprule
\textbf{Model} & \textbf{Base LLM} & \multicolumn{3}{c}{\textbf{GPT-4o Score} ($\uparrow$)} & \textbf{UTMOS} ($\uparrow$) & \textbf{WER} ($\downarrow$) & \textbf{Latency} ($\downarrow$) \\
 &  & \textit{General QA} & \textit{Knowledge} & \textit{Avg.} & (1-5) & (\%) & (ms) \\
\midrule
Whisper+LLM+XTTS  & LLaMA 3.1 8B & 6.70 & 7.70 & 7.20 & 4.23 & 1.70 & ~4200 \\
\midrule
SpeechGPT & LLaMA 2 13B & 1.40 & 2.20 & 1.80 & 3.86 & 66.57 & ~4000 \\
Mini-Omni & Qwen2 0.5B & 2.7 & 2.4 & 2.55 & 3.24 & 26.12 & ~350 \\ 
Llama-Omni & LLaMA 3.1 8B & 3.44 & 3.84 & 3.64 & 3.32 & 9.18 & ~220 \\
Moshi & Helium 7B & 2.71 & 3.91 & 3.31 & 3.92 & 7.97 & ~320 \\
GLM-4-Voice & GLM-4 9B & 5.24 & 5.67 & 5.30 & 3.97 & 6.40 & ~2500 \\
Freeze-Omni & Qwen2 7B & 3.48 & 4.98 & 4.23 & \textbf{4.38} & 14.05 & ~340 \\
MiniCPM-o 2.6 & Qwen2.5 7B & 5.46 & 6.21 & 5.84 & 3.87 & 10.60 & ~1200 \\
\midrule
\textbf{Whisper+LLM+LLMVoX (Ours)} & LLaMA 3.1 8B & \textbf{6.14} & \textbf{7.62} & \textbf{6.88} & 4.05 & \textbf{3.70} & ~475 \\
\bottomrule
\end{tabular}
\caption{Performance comparison of our framework (Whisper+LLM+LLMVoX) with other streaming speech-enabled LLMs and cascaded systems. Our system, which integrates \textbf{Whisper Small (224M)} for ASR and \textbf{LLMVoX (30M)} for text generation, achieves superior QA capabilities (6.14/7.62) compared to fine-tuned speech-enabled LLMs, while maintaining competitive speech quality (UTMOS 4.05) and low latency (475ms). Our model demonstrates superior text-speech alignment with a WER of 3.70\%.}
\label{tab:main_results}
\end{table*}

\subsection{Evaluation Tasks} We evaluate LLMVoX on five key tasks: \textbf{General QA Capability} assesses the model's ability to generate coherent and informative responses to general queries, reflecting the preservation of the LLM’s reasoning; \textbf{Knowledge Retention} measures the accuracy on fact-based questions to ensure robust information; \textbf{Speech Quality} examines the naturalness and clarity of the generated speech; \textbf{Speech-Text Alignment} verifies the consistency between the synthesized speech and corresponding text generated by the LLM. 
\textbf{Latency} is defined as the total elapsed time from when a query is submitted to when the model begins speaking.

\subsection{Evaluation Datasets and Baselines} \paragraph{Datasets.} We evaluate LLMVoX using diverse datasets spanning multiple dimensions. For \textbf{General QA}, we use questions from the AlpacaEval helpful-base and Vicuna subset \cite{alpaca_eval}, excluding math-related queries. For \textbf{Knowledge QA}, 100 fact-based questions are sourced from Web Questions \cite{berant2013semantic} and TriviaQA-verified \cite{joshi2017triviaqa}. To assess multilingual adaptability, we synthesize approximately 1,000 \textbf{Arabic} sentences from various domains. Additionally, for \textbf{Chunk Size Analysis}, we synthesize around 1,000 English sentences covering various topics, benchmarking the effects of chunk size on WER, CER, UTMOS, and latency.
We also evaluate on Visual Speech Question Answering task (VSQA) on LLaVA-Bench (In-the-Wild) \cite{liu2024mmbench}, which consists of 24 diverse images and 60 open-ended questions spanning various domains that suit conversational systems. We convert the text question to speech queries using \mbox{XTTS ~\cite{casanova2024xtts}.} 
\paragraph{Comparison Models.} 
LLMVoX is compared against recent speech-enabled LLMs:
\textbf{SpeechGPT}~\cite{zhang2023speechgpt} (7B, expanded vocabulary), 
\textbf{Mini-Omni}~\cite{xie2024mini} (0.5B, trained on VoiceAssistant-400K), 
\textbf{Llama-Omni}~\cite{fang2024llama} (LLaMA-3.1-8B with CTC speech head), 
\textbf{Moshi}~\cite{defossez2024moshi} (7B Helium model, dual-channel processing), 
\textbf{GLM-4-Voice}~\cite{zeng2024glm} (9B bilingual model with ultra-low bitrate tokenizer), and 
\textbf{Freeze-Omni}~\cite{wang2024freeze} (7B model with frozen LLM core) and \textbf{MiniCPM-o 2.6 }\cite{minicpm}. We also benchmark a cascaded pipeline with non-streaming TTS like XTTS\cite{casanova2024xtts}. All the models were evaluated on the basis of the best configuration given in the paper or the default configuration in the codebase. For Arabic TTS, no streaming comparison exists; hence we compare to non-streaming models - XTTS\cite{casanova2024xtts}, ArTST \cite{toyin2023artst}, FastPitch \cite{lancucki2021fastpitch}, Tacotron~2 \cite{elias2021parallel} and \mbox{Seamless \cite{barrault2023seamless} in Table \ref{tab:arabic_performance}.}

\subsection{Evaluation Protocol}
\textbf{General QA and Knowledge Tasks:}
The questions are first converted into speech using XTTS with multiple speaker modes to introduce input variation. Model streaming speech responses are saved and transcribed using \textbf{Whisper-Large-v3} \cite{radford2023robust}, and GPT-4o evaluates the quality and correctness of these transcriptions. For \textbf{General QA}, responses are scored from 1 to 10 based on coherence, informativeness, and fluency, following \textbf{MT-Bench protocols} \cite{zheng2023judging}. For \textbf{Knowledge QA}, \textbf{GPT-4o} compares responses against ground-truth answers, with scores 0 for incorrect and 1 for correct response. The total accuracy score is then normalized from 1 to 10.
Details of the evaluation prompts are \mbox{given in Appendix~\ref{prompts}.}

 \noindent \textbf{Speech Quality:} Naturalness is assessed using \textbf{UTMOS} \cite{saeki2022utmos}, predicting MOS \mbox{scores on a 1-5 scale.}

 \noindent \textbf{Speech-Text Alignment:} ASR Word Error Rate (WER) is calculated by comparing \textbf{Whisper-Large-v3} \cite{radford2023robust} transcriptions of the speech outputs with the LLM generated text averaged over General and Knowledge QA tasks.

 \noindent \textbf{Latency:} Measured from the reception of speech input to the first speech output, capturing both processing and synthesis delays.

\noindent \textbf{Human Evaluation:} We compare our system with \textbf{Freeze-Omni}, one of the
closely related approaches that freeze the base LLM. For setup details, see Appendix~\ref{app:human_eval}.

\begin{figure} [!t]
    \centering
    \includegraphics[width=\linewidth]{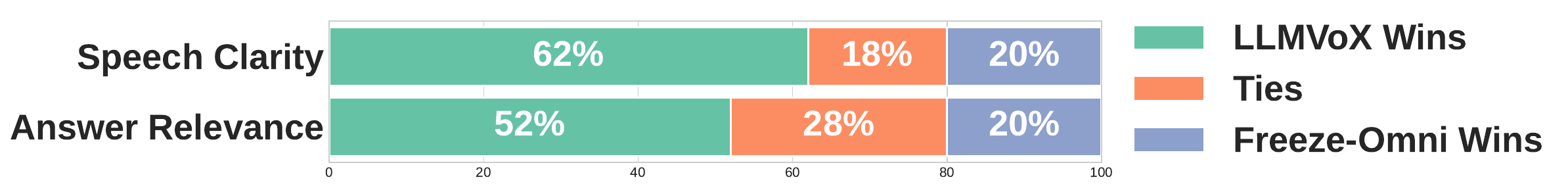}
    \caption{Human evaluation:  Comparing with Freeze-Omni on Answer Relevance and Speech Quality.}
    \label{fig:human_eval_chart.pdf}
\end{figure}

\subsection{Experimental Results}
\label{subsec:results}
\noindent \textbf{Linguistic Capabilities:} Our modular setup with Whisper for ASR, LLama 3.1 8B \cite{dubey2024llama} and LLMVoX achieves the highest GPT-4o score (see Table~\ref{tab:main_results}) among streaming models with 6.14 (General QA) and 7.62 (Knowledge QA) demonstrating its ability to preserve LLaMA 3.2 8B’s language understanding capabilities. Although XTTS slightly outperforms LLMVoX sharing the same base LLM due to lower WER, its high latency (4200ms vs 475ms) makes it impractical for real-time use, highlighting the efficiency of LLMVoX. Notably, LLaMA-Omni, despite using the same LLaMA 3.1 8B base, underperforms in both QA tasks (3.44 vs. 6.14, 3.84 vs. 7.62), suggesting LLM degradation. Similarly, Freeze-Omni, despite freezing its LLM backbone, suffers from a high WER (14.05\%), which lowers coherence and response quality. Also, based on human evaluation results in Figure ~\ref{fig:human_eval_chart.pdf}, we observe that the response quality of our framework is much better than similar approach like Freeze-Omni that also its LLM parameters frozen.

\noindent \textbf{Speech Quality \& Alignment:} 
While Freeze-Omni yields a high UTMOS (Table~\ref{tab:main_results}), its WER is substantially high (14.05\%), indicating a misalignment between the generated speech and text. In contrast, LLMVoX achieves the lowest WER at 3.70\%, demonstrating superior text-to-speech consistency while maintaining a strong UTMOS score of 4.05. From the human evaluation results in Figure~\ref{fig:human_eval_chart.pdf}, our approach favours speech clarity compared to Freeze-Omni by a significant margin.

\begin{figure}[t]
    \centering
    \includegraphics[width=0.8\linewidth,trim={0 3mm 0 0},clip]{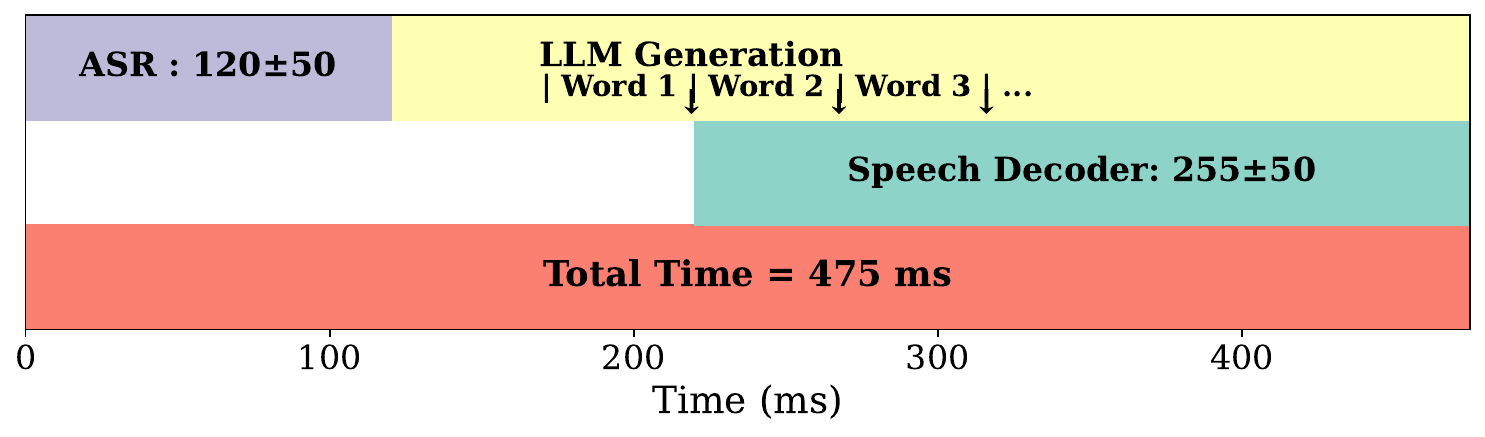}
    \caption{Breakdown of average end-to-end latency (in milliseconds) at a chunk size of 40 for a single query.}
    \label{tab:latency_breakdown}
\end{figure}

\noindent \textbf{Latency Analysis:}
One of the key challenges in real-time TTS is balancing low latency with high speech quality. LLMVoX successfully achieves this, delivering an end-to-end latency of 475ms, making it competitive with end-to-end streaming-capable models while significantly improving upon cascaded approaches like Whisper+LLM+XTTS (4200ms). While Llama-Omni achieves lower latency (220ms), its trade-off in WER (9.18\%) and low UTMOS score of 3.32. In contrast, LLMVoX achieves a more optimal balance, reducing latency by nearly 86\% compared to XTTS while maintaining superior WER. This is crucial for applications where both real-time response and textual accuracy are equally important, such as voice assistants. Figure~\ref{tab:latency_breakdown} shows that LLMVoX starts generating speech tokens the moment LLM generates the first word, unlike other chain-of-modality models and cascaded pipelines, to achieve very low latency while operating in parallel to the LLM.

\begin{figure}[!t]
    \centering
    \setlength{\tabcolsep}{0pt}
   \begin{tabular}{cc}
        \includegraphics[width=0.5\columnwidth]{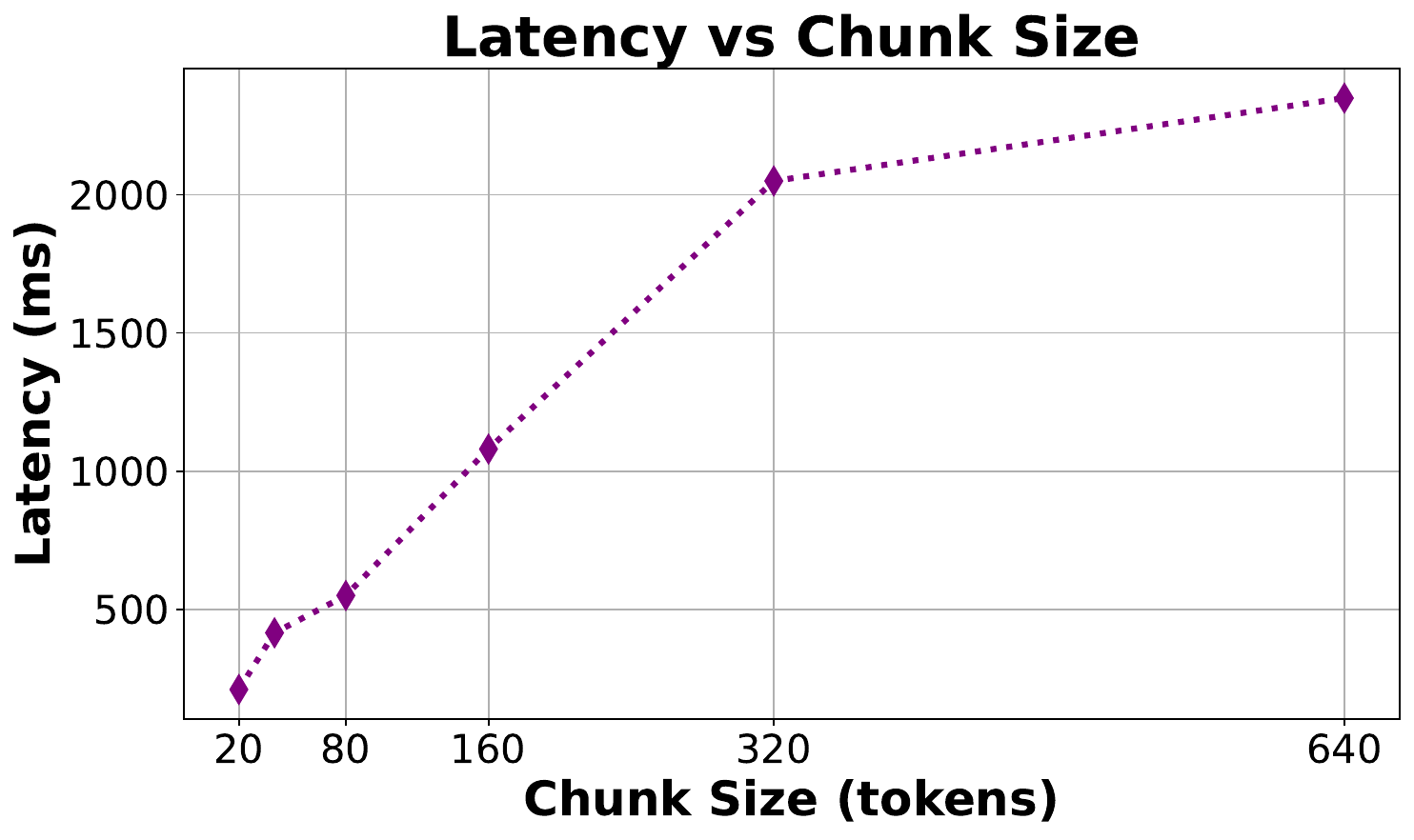}  &
        \includegraphics[width=0.5\columnwidth]{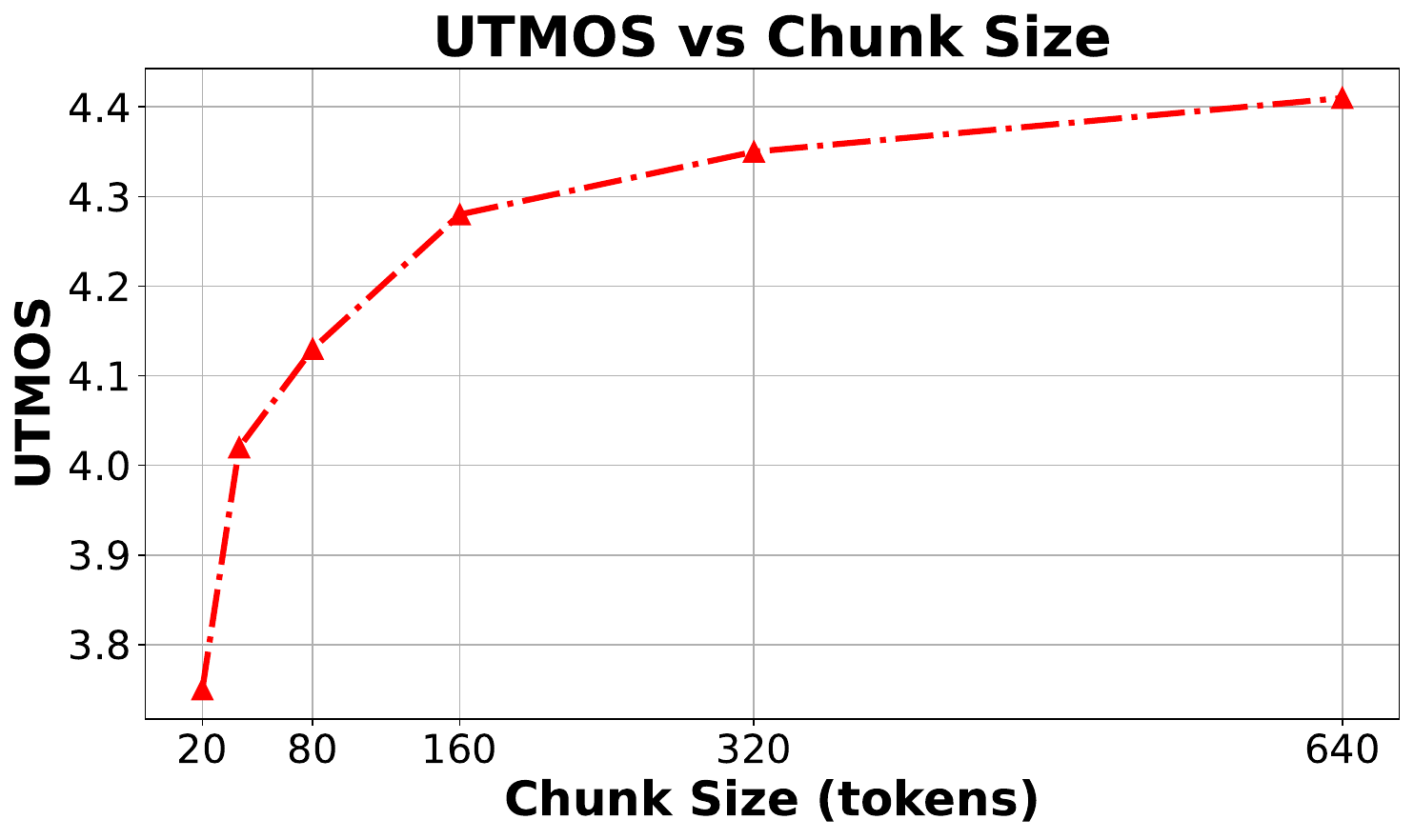} \\
    
        \includegraphics[width=0.5\columnwidth,trim={0 5mm 0 0},clip]{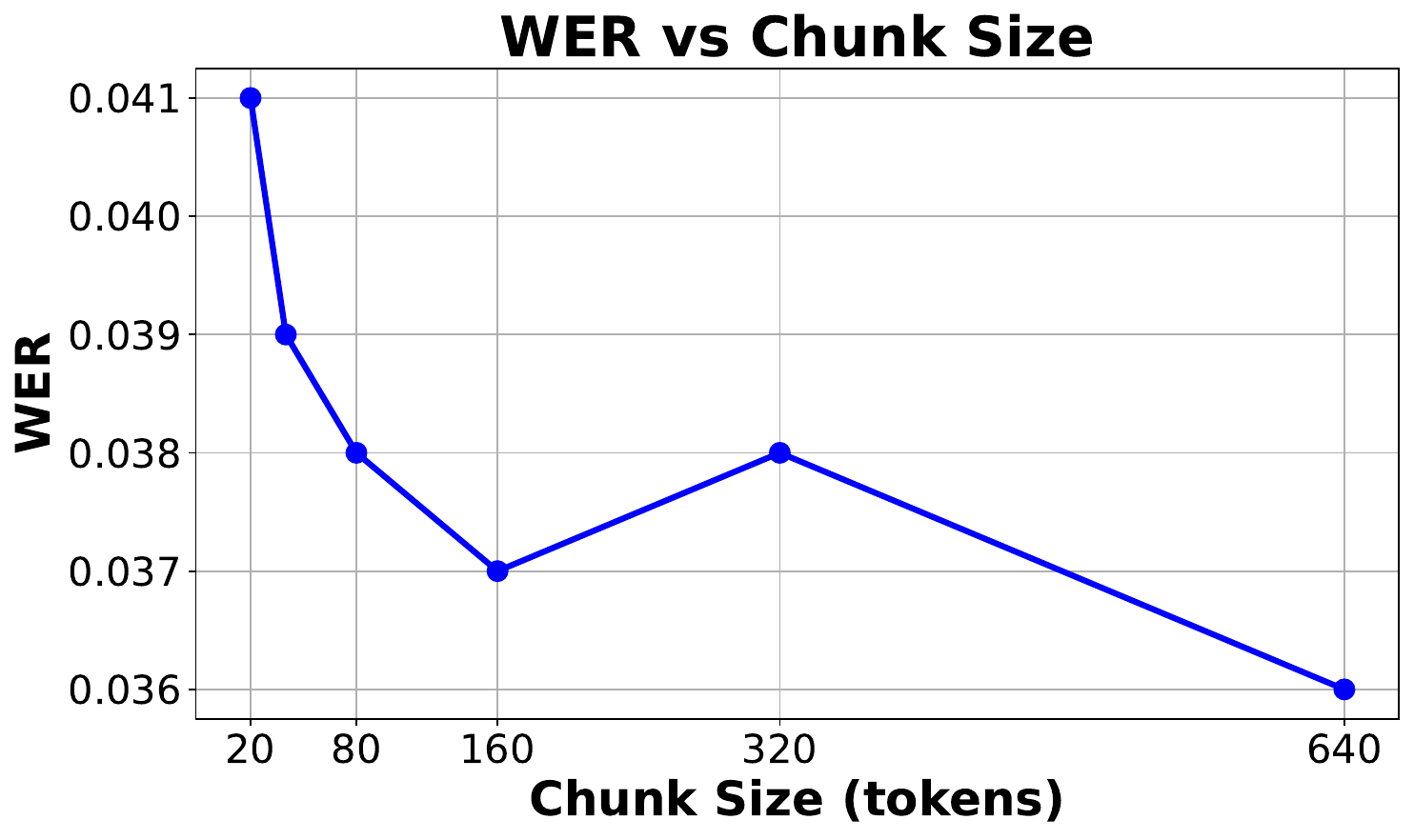}  &   
        \includegraphics[width=0.5\columnwidth,trim={0 5mm 0 0},clip]{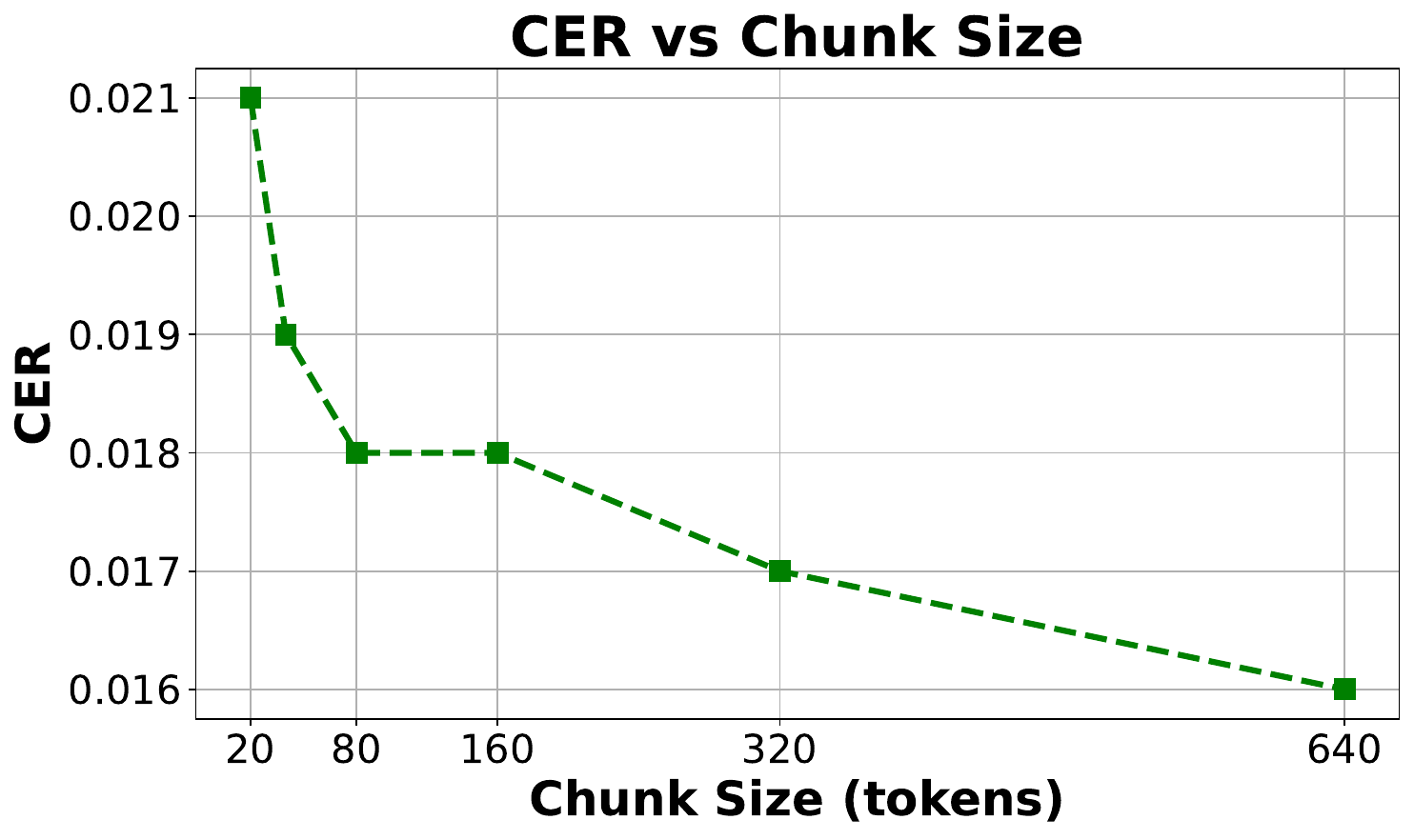} 
    \end{tabular}
    \caption{Effect of chunk size on WER, CER, UTMOS, and latency. Larger chunks enhance speech quality and reduce error rates.}
    \label{fig:chunk_analysis}
\end{figure}

\begin{table}[!t]
\centering
\small  
\setlength{\tabcolsep}{2pt}  
\scalebox{0.9}{
\begin{tabular}{lcc}
\toprule
\textbf{LLM} & \textbf{Params} & \textbf{Latency (s)} \\
\midrule
Qwen 2.5      & 0.5B  & 0.33 \\
Lamma 3.2     & 3B    & 0.36 \\
Lamma 3.1     & 8B    & 0.47 \\
Phi 4         & 14B   & 0.95 \\
Mixtral Small & 24B   & 1.25 \\
Qwen 2.5      & 32B   & 1.40 \\
Lamma 3.3     & 70B   & 1.91 \\
\bottomrule
\end{tabular}
}
\caption{End-to-end latency(ASR included) of LLMVoX with various LLMs at chunk size of 40.}
\label{tab:latency_llm}
\end{table}

\begin{table}[!t]
\centering
\resizebox{\columnwidth}{!}{%
\begin{tabular}{lccc}
\toprule
\textbf{Model} & \textbf{Streaming} & \textbf{WER} ($\downarrow$) & \textbf{CER} ($\downarrow$) \\
\midrule
XTTS                         & No  & \textbf{0.062} & \textbf{0.017} \\
ArTST                        & No  & 0.264 & 0.125 \\
FastPitch Arabic Finetuned   & No  & 0.493 & 0.153 \\
Tacotron 2 Arabic Finetuned  & No  & 0.663 & 0.268 \\
Tacotron 2 Arabic Finetuned  & No  & 0.663 & 0.268 \\
Seamless-M4t-Large           & No  & 0.342 & 0.145 \\
\textbf{LLMVoX (Ours)} & Yes & 0.234 & 0.082 \\
\bottomrule
\end{tabular}
}
\caption{Arabic TTS performance comparison. LLMVoX achieves competitive error rates in a streaming setup,  operating at nearly 10x faster speed compared to state-of-the-art XTTS.}
\label{tab:arabic_performance}
\end{table}

\begin{table}[!t]
\centering
\renewcommand{\arraystretch}{1} 
\setlength{\tabcolsep}{3pt} 
\resizebox{0.95\columnwidth}{!}{ 
\begin{tabular}{@{}lcccc@{}}
\toprule
\textbf{Model} & \textbf{WER} & \textbf{CER} & \textbf{GPT Score} & \textbf{Latency (s)} \\
\midrule
MiniCPM-o 2.6 & 0.053 & 0.036 & 6.32 & 1.45 \\
\textbf{LLMVoX (Ours)} & 0.042 & 0.022 & 6.41 & 1.05 \\
\bottomrule
\end{tabular}}
\caption{VSQA performance on LLaVA-Bench (In-the-Wild) with Qwen 2.5 VL 7B as the backbone.}
\label{tab:vlqa_results}
\end{table}

\noindent \textbf{Observations on Chunk Size Impact:}
From Figure~\ref{fig:chunk_analysis}, we see that increasing the initial chunk size improves overall synthesis quality without significantly increasing latency. Key observations include:
\textbf{UTMOS} improves from 3.75 to 4.41 as chunk size increases, suggesting speech reconstruction from larger chunk size results in smoother and more natural prosody.
\textbf{WER} decreases from 0.041 to 0.036 confirming that larger chunks improve phonetic consistency.
Latency remains under 1 second for chunk sizes as large as 160 ensuring real-time usability despite larger chunk sizes.

\noindent \textbf{Latency Analysis with LLM Integration}
Table~\ref{tab:latency_llm} shows that LLMVoX latency at a chunk size of 40 increases with LLM size. Smaller models like Qwen 2.5 (0.5B) and Lamma 3.2 (3B) achieve lower latencies (0.33–0.36s), while larger models such as Phi 4 (14B) and Lamma 3.3 (70B) exceed 1s. This indicates that while larger LLMs impose higher computational costs, architectural optimizations also impact latency.

\subsection{Arabic Multilingual Performance:} On the curated Arabic eval set, LLMVoX achieves a CER of 8.2\%, outperforming most non-streaming TTS methods except XTTS which was used to synthesize the Arabic Training data suggesting robust adaptability to new languages without explicit Grapheme-to-Phone(G2P) conversion or training. 

\subsection{Adaptability with Vision language Models}
To demonstrate our method's versatility, we integrate LLMVoX into a multimodal pipeline for Visual Speech Question Answering (VSQA). Our setup combines \textbf{Whisper-Small} for ASR, \textbf{Qwen 2.5-VL-7B}~\cite{qwen2.5-VL} for visual-language processing, and LLMVoX for speech synthesis.
Table~\ref{tab:vlqa_results} compares our system with the omni-multimodal MiniCPM-o 2.6 model\cite{minicpm}. We report word error rate (WER), character error rate (CER), and GPT-4o score.
Our system achieves lower WER and a comparable GPT score, demonstrating that LLMVoX can be effectively plugged into state-of-the-art VLM pipelines for \mbox{challenging speech VQA tasks.}

\section{Conclusion}
We introduce LLMVoX, an LLM-agnostic autoregressive streaming TTS that decouples speech synthesis from text generation. Leveraging a lightweight Transformer and multi-queue streaming, LLMVoX delivers high-quality, continuous speech with minimal latency while preserving LLM reasoning. Experiments on English and Arabic tasks show that LLMVoX outperforms or matches other speech-enabled LLMs, offering a scalable solution for real-time multimodal AI.

\section{Limitations}
LLMVoX achieves low-latency streaming TTS without modifying the underlying LLM, but it has the following limitations. First, the system lacks voice cloning, which limits its ability to generate speaker-specific vocal characteristics—a key feature for personalized interactions. Second, while we use Whisper for ASR, it is not fully integrated into the streaming pipeline, leaving potential latency reductions unexplored. Future work will focus on incorporating voice cloning and extending the streaming architecture to the ASR input, further enhancing personalization and \mbox{real-time performance.}



\bibliography{main}

\clearpage

\section{Appendix}
\subsection{Prompt for Evaluating Spoken Chatbots}
\label{prompts}
This section describes the two primary GPT-4o prompts we use for evaluating spoken chatbot responses. Each prompt targets a different aspect of performance: (1) the overall quality of an answer (General QA) and (2) the correctness of the answer compared to reference responses (Knowledge).

\subsubsection{General QA}

\noindent
\textbf{[Instruction]}\\
Please act as an impartial judge and evaluate the quality of the response provided by an AI assistant to the user question displayed below. Your evaluation should consider factors such as the helpfulness, relevance, accuracy, depth, creativity, and level of detail of the response. Begin your evaluation by providing a short explanation. Be as objective as possible. After providing your explanation, you must rate the response on a scale of 1 to 10 by strictly following this format: ``Rating: [[5]]''.

\noindent
\textbf{[Question]}\\
\textit{\{User's question goes here\}}

\noindent
\textbf{[The Start of Assistant’s Answer]}\\
\textit{\{Assistant's response begins here\}}\\
\textbf{[The End of Assistant’s Answer]}

\subsubsection{Knowledge}

\noindent
\textbf{[Instruction]}\\
You will be given a question, the reference answers to that question, and an answer to be judged. Your task is to judge whether the answer to be judged is correct, given the question and reference answers. An answer is considered correct if it expresses the same meaning as at least one of \mbox{the reference answers.}

\noindent You should respond in JSON format. First provide a concise one-sentence analysis in the field ``analysis'', then your final judgment in the field ``judgment'', which can be ``correct'' or ``incorrect''.

\noindent
\textbf{[Question]}\\
\textit{\{User's question\}}

\noindent
\textbf{[Reference Answer]}\\
\textit{\{targets\}}

\noindent
\textbf{[Answer To Be Judged]}\\
\textit{\{answer\_to\_be\_judged\}}

\noindent 
\textbf{Example Output (in JSON format)}:

\begin{Verbatim}[frame=single]
{
  "analysis": "A concise explanation of 
        correctness or incorrectness.",
  "judgment": "correct"
}
\end{Verbatim}


\noindent
These prompts enable both qualitative (General QA) and correctness-based (Knowledge) evaluations of AI-generated spoken responses, ensuring a comprehensive assessment of \mbox{the system's performance.}
\subsection{Human Evaluation Setup and Conclusion}
\label{app:human_eval}
We conducted a human evaluation to compare the streaming speech outputs of our proposed system with those of \textbf{Freeze-Omni}. Specifically, we randomly selected 30 questions from various domains and generated responses using both systems. These responses were distributed in batches of five per user, with a total of 20 users participating in the evaluation. For our system, we use Whisper-Small for ASR, LLaMA 3.1 8B as the LLM, and LLMVoX for streaming TTS, while \textbf{Freeze-Omni} served as the baseline.
\noindent
The streaming speech responses were recorded and a custom user interface was developed to facilitate evaluation. Participants listened to each response and rated the best response based on two metrics:
\\
\textbf{(i)Answer Relevance}: Evaluates how factual, useful, and relevant the answer is to the question.
\\
\textbf{(ii)Speech Quality}: Assesses the flow, word clarity, and pronunciation of the generated speech.\\
These choices were then aggregated to compare the overall performance of the two systems. The aggregated results are illustrated in Figure~\ref{fig:human_eval_chart.pdf}
Our human evaluation results indicate that our proposed system outperforms Freeze-Omni on both key metrics. Based on responses to the 30 questions, LLMVoX integrated with Whisper-Small for ASR and LLaMA 3.1 8B as the LLM received higher user ratings for both answer relevance and speech quality. Specifically, our model achieved wins in 52\% of cases for answer relevance and 62\% for speech quality, compared to Freeze-Omni’s 20\% wins on each metric. These findings suggest that decoupling speech synthesis from text generation not only preserves the linguistic capabilities of the LLM but also produces more natural, clear, and engaging speech output, demonstrating the effectiveness of our approach for \mbox{real-time dialogue applications.}

\end{document}